\RequirePackage{fix-cm}
\documentclass[twocolumn]{svjour3}          
\smartqed  
\usepackage{graphicx}
\usepackage{ae}
\usepackage{comment}
\usepackage{float}
\usepackage{amsmath,url} 
\usepackage{amssymb}
\usepackage{color}
\usepackage{bm}
\usepackage{dirtytalk}
\usepackage[dvipsnames]{xcolor}
\usepackage{tensor}
\usepackage{braket} 
\DeclareMathOperator*{\E}{{\rm I\kern-.3em E}} 
 
\usepackage{multirow}
\usepackage[normalem]{ulem}
\usepackage{physics}
\usepackage{mathtools}
\usepackage{optidef}
\usepackage[ruled,vlined, linesnumbered]{algorithm2e}
\SetKwComment{Comment}{$\triangleright$\ }{}

\usepackage{multicol}
\usepackage{array}
\usepackage{booktabs}
\usepackage{longtable}
\usepackage{lscape}
\usepackage{pdflscape}

\usepackage{subcaption} 

\usepackage{titlesec}

\titleformat{\paragraph}
{\normalfont\normalsize\bfseries}{\theparagraph}{1em}{}
\titlespacing*{\paragraph}
{0pt}{3.25ex plus 1ex minus .2ex}{1.5ex plus .2ex}

{ }
 { }
 { }
 
 \newtheorem{exa}{{\bf Example}}{}
{}
\begin{document}

\title{Meta-Model Structure Selection: Building Polynomial NARX Model for Regression and Classification
}


\author{Wilson Rocha Lacerda Junior \and
        Samir Angelo Milani Martins \and
        Erivelton Geraldo Nepomuceno 
}


\institute{Wilson Rocha Lacerda Junior \at
              Control and Modelling and Control Group (GCOM),
              Department of Electrical Engineering, Federal University of S\~ao Jo\~ao del-Rei, Minas Gerais, Brazil \\
              \email{wilsonrljr@outlook.com}           
           \and
           Samir Angelo Milani Martins \at
              Control and Modelling and Control Group (GCOM),
              Department of Electrical Engineering, Federal University of S\~ao Jo\~ao del-Rei, Minas Gerais, Brazil \\
              \email{martins@ufsj.edu.br}
            \and
            Erivelton Geraldo Nepomuceno \at
            Control and Modelling and Control Group (GCOM),
            Department of Electrical Engineering, Federal University of S\~ao Jo\~ao del-Rei, Minas Gerais, Brazil \\
            \email{nepomuceno@ufsj.edu.br}
}

\date{Received: date / Accepted: date}

\maketitle

\begin{abstract}
This work presents a new meta-heuristic approach to select the structure of polynomial NARX models for regression and classification problems. The method takes into account the complexity of the model and the contribution of each term to build parsimonious models by proposing a new cost function formulation. The robustness of the new algorithm is tested on several simulated and experimental system with different nonlinear characteristics. The obtained results show that the proposed algorithm is capable of identifying the correct model, for cases where the proper model structure is known, and determine parsimonious models for experimental data even for those systems for which traditional and contemporary methods habitually fails. The new algorithm is validated over classical methods such as the FROLS and recent randomized approaches.
\end{abstract}

\section{Introduction}
\label{intro}
System identification is a method of identifying the dynamic model of a system from measurements of the system inputs and outputs ~\cite{Bil2013}. In particular, the interest in nonlinear system identification has deserved much attention by researchers from the $1950$s onward and many relevant results were developed \cite{Wie1958,Rug1981,HK1999,PS2012}. In this context, one frequently employed model representation is the NARMAX (Non-linear Autoregressive Models with Moving Average and Exogenous Input), which was introduced in $1981$ aiming at representing a broad class of nonlinear system \cite{BL1981,LB1985,CB1989}.

There are many NARMAX model set representations such as polynomial, generalized additive, and neural networks. Among these types of the extended model set, the power-form polynomial is the most commonly NARMAX representation~\cite{Bil2013}. Fitting polynomial NARMAX models is a simple task if the terms in the model are known \emph{a priori}, which is not the case in real-world problems. Selecting the model terms, however, is fundamental if the goal of the identification is to obtain models that can reproduce the dynamics of the original system. Problems related to overparameterization and numerical ill-conditioning are typical because of the limitations of the identification algorithms in selecting the appropriate terms that should compose the final model ~\cite{AB1995,PS2003}.  

In that respect, one of the most traditionally algorithms for structure selection of polynomial NARMAX was developed by ~\cite{KBLM1988} based on the Orthogonal Least Squares (OLS) and the Error Reduction Ratio (ERR), called Forward Regression Orthogonal Least Squares (FROLS). Numerous variants of FROLS algorithm has been developed to improve the model selection performance such as~\cite{BCK1989,FP2012,GGBW2015,MB1999}. The drawbacks of the FROLS have been extensively reviewed in the literature, \emph{e.g.}, in~\cite{BA1995,PP2001,FPP2015}. Most of these weak points are related to i) the Prediction Error Minimization (PEM) framework; ii) the inadequacy of the ERR index in measuring the absolute importance of regressors; iii) the use of information criteria such as Akaike Information Criterion (AIC) ~\cite{Aka1974}, Final Prediction Error (FPE) ~\cite{Aka1969} and the Bayesian information criterion (BIC) ~\cite{Sch1978}, to select the model order. Regarding the information criteria, although these techniques work well for linear models, in a nonlinear context no simple relation between model size and accuracy can be established ~\cite{FPP2015,CHH2003}.

As a consequence of the limitations of OLS based algorithms, some recent research endeavors have significantly strayed from the classical FROLS scheme, by reformulating the Model Structure Selection (MSS) process in a probabilistic framework and using random sampling methods \cite{FPP2015,TCD2012,BAK2013,RFF2004,SA2017}. Nevertheless, these techniques based on meta-heuristics and probabilistic frameworks 
presents some flaws. The meta-heuristics approaches turn on AIC, FPE, BIC and others information criteria to formulate the cost function of the optimization problem, generally resulting in over-parameterized models.

Last but not last, due to the importance of classification techniques for decision-making tasks in engineering, business, health science, and many others fields, it is surprising how only a few researchers have addressed this problem using classical regression techniques. The authors in~\cite{SWB2019} presented a novel algorithm that combines logistic regression with the NARX methodology to deal with systems with a dichotomous response variable. The results in that work, although very interesting, are based on FROLS algorithm and, therefore, inherits most of the drawbacks concerning the traditional technique, opening new paths for research.

This work proposes a technique to the identification of nonlinear systems using meta-heuristics that fills the mentioned gaps in what concerns the structure selection of NARMAX models for regression and classification. The method uses an alternative to the information criteria cited as the index indicating the accuracy of the model as a function of the size of the model. Finally, the proposed algorithm is adapted to deal with classification problems to represent systems with binary outputs that depend on continuous time predictors.

The remainder of this work is organized as follows: Section 2 provides the basic framework and notation for nonlinear system identification of NARX models. Section 3 presents the necessary tools to formulate the cost function of the identification strategy. This section also introduces the new algorithm and reports the results obtained on several systems taken from the literature and physical systems. Section 4 adapts the technique to develop NARX models considering systems with binary responses that depend on continuous predictors. Section 5 recaps the primary considerations of this study and proposes possible future works.

\section{Background}
\subsection{Polynomial NARX model}
Polynomial Multiple-Input Multiple-Output (MIMO) NARX is a mathematical model based on difference equations and relates the current output as a function of past inputs and outputs, mathematically described as~\cite{BCK1989,LB1985}:
\begin{align}
\label{eq5:narx_mimo_general}
             y_{\indices{_i}k}=& F_{\indices{_i}}^\ell \bigl[y_{\indices{_1}k-1},  \dotsc, y_{\indices{_1}k-n^i_{y\indices{_1}}},\dotsc, y_{\indices{_s}k-1},  \dotsc, y_{\indices{_s}k-n^i_{y\indices{_s}}}, \nonumber \\
             & x_{\indices{_1}k-d}, x_{\indices{_1}k-d-1}, \dotsc, x_{\indices{_1}k-d-n^i_{x\indices{_1}}}, \dotsc, \nonumber \\
             &x_{\indices{_r}k-d}, x_{\indices{_r}k-d-1}, \dotsc, x_{\indices{_r}k-d-n^i_{x\indices{_r}}}\bigr] + \xi_{\indices{_i}k},
\end{align}
\noindent where $n_y \in \mathbb{N}^*$, $n_x \in \mathbb{N}$, are the maximum lags for the system output and input respectively; $x_k \in \mathbb{R}^{n_x}$ is the system input and $y_k\in \mathbb{R}^{n_y}$ is the system output at discrete time~$k \in \mathbb{N}^n$; $e_k \in \mathbb{R}^{n_e}$ stands for uncertainties and possible noise at discrete time~$k$. In this case,$\mathcal{F}^\ell$ is some nonlinear function of the input and output regressors with nonlinearity degree $\ell \in \mathbb{N}$ and~$d$ is a time delay typically set to~$d~=~1$. 

The number of possibles terms of MIMO NARX model given the $i$th polynomial degree, $\ell_i$, is:
\begin{equation}
\label{eq5:nr_mimo}
    n_{\indices{_{m}}r} = \sum_{j = 0}^{\ell_i}n_{ij},
\end{equation}
\noindent where
\begin{align}
    n_{ij} = \frac{ n_{ij-1} \biggl[ \sum\limits_{k=1}^{s} n^i_{y_k} + \sum\limits_{k=1}^{r} n^i_{x_k} + j - 1 \biggr]}{j}, \nonumber \\ \qquad n_{i0}=1, j=1, \dotsc, \ell_i.
\label{eq5:nr_mimo1}
\end{align}

Parsimony makes the Polynomial NARX models a widely known model family. This characteristic means that a wide range of behaviors can be represented concisely using only a few terms of the vast search space formed by candidate regressors and usually a small data set are required to estimate a model.

\subsection{Importance of Structure Selection}
Identifying the correct structure, is fundamental to allow the user to be able to analyze the system dynamics consistently. The regressors selection, however, is not a simple task. If $\ell$, $n_x$, and $n_y$, increases, the number of candidate models becomes too large for brute force approach. Considering the MIMO case, this problem is far worse than the Single-Input Single-Output (SISO) one if many inputs and outputs are required. The total number of all different models is given by
\begin{align}
    n_m = 
    \begin{cases}
    2^{n_r} & \text{for SISO models}, \\
    2^{n_{\indices{_{m}}r}} & \text{for MIMO models},
    \end{cases}
\end{align}

\noindent where $n_r$ and $n_{\indices{_{m}}r}$ are the values computed using Eq.~\eqref{eq5:nr_mimo} to Eq.~\eqref{eq5:nr_mimo1}.

A classical solution to regressors selection problem is the FROLS algorithm associated with ERR test. The FROLS method adapt the set of regressors in the search space into a set of orthogonal vectors, which ERR evaluates the individual contribution to the desired output variance by calculating the normalized energy coefficient $C(x,y)$ between two vectors defined as:
\begin{equation}
    C(x,y) = \frac{(x^\top y)^2}{(x^\top x)(y^\top y)}.
\end{equation}    

An approach often used is to stop the algorithm using some information criteria, \emph{e.g.}, AIC~\cite{Aka1974}.

\subsection{The Sigmoid Linear Unit Function}

\begin{definition}[Sigmoidal function]
\label{sigmoidal-function}
Let $\mathcal{F}$ represent a class of bounded functions $\phi: \mathbf{R} \mapsto \mathbf{R} $. If the properties of $\phi(x)$ satisfies 
\begin{align}
    &\lim\limits_{x \to \infty} \phi(x) = \alpha \nonumber \\
    &\lim\limits_{x \to -\infty} \phi(x) = \beta \quad \text{with } \alpha > \beta,  \nonumber
\end{align}
\noindent the function is called sigmoidal.
\end{definition}

In this particular case and following definition Eq.~\eqref{sigmoidal-function} with $alpha = 0 $ and $\beta = 1$, we write a "S" shaped curve as
\begin{equation}
    \label{s-shape}
    \varsigma(x) = \frac{1}{1+e^{-a(x-c)}}.
\end{equation}

In that case, we can specify $a$, the rate of change. If $a$ is close to zero, the sigmoid function will be gradual. If $a$ is large, the sigmoid function will have an abrupt or sharp transition. If $a$ is negative, the sigmoid will go from $1$ to zero. The parameter $c$ corresponds to the x value where $y = 0.5$.

The Sigmoid Linear Unit Function (SiLU) is defined by the sigmoid function multiplied by its input
\begin{equation}
    \label{silu-shape}
    \textsc{silu}(x) = x \varsigma(x),
\end{equation}
\noindent which can be viewed as an steeper sigmoid function with overshoot.

\subsection{Meta-heuristics}
In general, nature-inspired optimization algorithms have been increasingly widespread over the last two decade due to the flexibility, simplicity, versatility, and local optima avoidance of the algorithms in real life applications. 

Two essential characteristics of meta-heuristics algorithms are exploitation and exploration~\cite{BR2003}. Exploitation is related to the local information in the search process regarding the best near solution. On the other hand, exploration is related to explore a vast area of the search space to find an even better solution and not be stuck in local optima. \cite{EC1998} shows that there is no consensus about the notion of exploration and exploitation in evolutionary computing, and the definitions are not generally accepted. However, it can be observed a general agreement about they work like opposite forces and usually hard to balance. In this sense, a combination of two metaheuristics, called hybrid metaheuristic, can be done to provide a more robust algorithm.

\subsubsection{The Binary hybrid Particle Swarm Optimization and Gravitational Search Algorithm (BPSOGSA) algorithm}
As can be observed in most meta-heuristics algorithm, to achieve a good balance between exploration and exploitation phase is a challenging task. In this paper, to provide a more powerful performance by assuring higher flexibility in the search process a BPSOGSA hybridized using a low-level co-evolutionary heterogeneous technique~\cite{Tal2002} proposed by~\cite{MH2010} is used. The main concept of the BPSOGSA is to associate the high capability of the particles in Particle Swarm Optimization (PSO) to scan the whole search space to find the best global solution with the ability to look over local solutions of the Gravitational Search Algorithm (GSA) in a binary space.

\subsubsection{Standard PSO algorithm}
In PSO~\cite{KE1995,Ken2010}, each particle represents a candidate solution and consists of two parts: the location in the search space, $\vec{x}_{\,np,d} \in \mathbb{R}^{np\times d}$, and the respective velocity, $\vec{v}_{\, np, d} \in \mathbb{R}^{np \times d}$, where $np = 1,2, \cdots, n_a$ and $n_a$ is the size of the swarm and $d$ is the dimension of the problem. In this respect, the following equation represents the initial population:
\begin{align}
    \vec{x}_{\,np,d} = 
        \begin{bmatrix}
        x_{1,1} & x_{1,2} & \cdots & x_{1,d} \\
        x_{2,1} & x_{2,2} & \cdots & x_{2,d} \\
        \vdots  & \vdots  & \ddots & \vdots \\
        x_{n_a,1} & x_{n_a,2} & \cdots & x_{n_a,d}
        \end{bmatrix}
        \label{eq5:population}
\end{align}

At each iteration, $t$, the position and velocity of a particle are updated according to
\begin{align}
    v_{nv,d}^{t+1} = \zeta v_{nv,d}^{t} + c_1 \kappa_1 (pbest_{np}^{t} - x_{np,d}^{t}) \nonumber \\ 
    + c_2 \kappa_2 (gbest_{np}^{t} - x_{np,d}^{t}),
\label{eq5:velocity_pso}
\end{align}
\noindent where $\kappa_j \in \mathbb{R}$, for $j = [1,2]$, are a real-valued, continuous random variable in the interval $[0,1]$, $\zeta \in \mathbb{R}$ is an inertia factor to control the influence of the previous velocity on the current one (also working representing a trade-off between exploration and exploitation), $c_1$ is the cognitive factor related to $pbest$ (best particle) and $c_2$ is the social factor related to $gbest$ (global solution). The values of the velocity, $\vec{v}_{\, np,d}$, are usually bounded in the range $[v_{min}, v_{max}]$ to guarantee that the randomness of the system do not lead to particles rushing out of the search space. The position are updated in the search space according to
\begin{equation}
    x_{np,d}^{t+1} = x_{np,d}^{t} + v_{np,d}^{t+1},
\label{eq5:location_pso}
\end{equation}

\subsubsection{Standard GSA algorithm}
In GSA~\cite{RNS2009}, the agents are measured by their masses, which are proportional to their respective values of the fitness function. These agents share information related to their gravitational force in order to attract each other to locations closer to the global optimum. The larger the values of the masses, the best possible solution is achieved, and the agents move more slowly than lighter ones. In GSA, each mass (agent) has four specifications: position, inertial mass, active gravitational mass, and passive gravitational mass. The position of the mass corresponds to a solution to the problem, and its gravitational and inertial masses are determined using a fitness function. 

Consider a population formed by agents described in Eq.~\eqref{eq5:population}. At a specific time $t$, the velocity and position of each agent are updated, respectively, as follow:

\begin{align}
    v_{i,d}^{t+1} &= \kappa_i \times v_{i,d}^t + a_{i,d}^t, \nonumber\\
    x_{i,d}^{t+1} &= x_{i,d}^t +v_{i,d}^{t+1}.
    \label{eq5:vel_pos_gsa}
\end{align}
\noindent where $\kappa$ gives a stochastic characteristic to the search. The acceleration, $a_{i,d}^t$, is computed according to the law of motion~\cite{RNS2009}:
\begin{equation}
    a_{i,d}^t = \frac{F_{i,d}^t}{M_{ii}^{t}},
\label{eq5:accelaration}
\end{equation}
\noindent where $t$ is a specific time, $M_{ii}$ is inertial the mass of object $i$ and $F_{i,d}$ the gravitational force acting on mass $i$ in a $d-$dimensional space. The detailed process to calculate and update both $F_{i,d}$ and $M_{ii}$ can be found in~\cite{RNS2009}.

\subsubsection{The binary hybrid optimization algorithm}
The combination of the algorithms are according to~\cite{MH2010}: 
\begin{align}
\label{eq5:vel_psogsa}
    v_{i}^{t+1} = \zeta \times v_i^t + \mathrm{c}'_{1} \times \kappa \times a_i^t + \mathrm{c}'_2 \times \kappa \times (gbest -x_i^t),
\end{align}
\noindent where $\mathrm{c}'_j \in \mathbb{R}$ is an acceleration coefficient. The Eq.~\eqref{eq5:vel_psogsa} have the advantage to accelerate the exploitation phase by saving and using the location of the best mass found so far. However, because this method can affect the exploration phase as well, \cite{ML2014} proposed a solution to solve this issue by setting adaptive values for $\mathrm{c}_j'$, described by~\cite{MWC2014}:
\begin{align}
    \mathrm{c}_1' &= -2\times\frac{t^3}{\max(t)^3}+2 \\
    \mathrm{c}_2' &= 2\times\frac{t^3}{\max(t)^3}+2 \\.
    \label{eq5:adapt_c}
\end{align}

In each iteration, the positions of particles are updated as stated in Eq.~\eqref{eq5:location_pso} to Eq.~\eqref{eq5:vel_pos_gsa}.

To avoid convergence to local optimum when mapping the continuous space to discrete solutions, the following transfer function are used~\cite{ML2013}:
\begin{equation}
    S(v_{ik}) = \abs{\frac{2}{\pi}\arctan(\frac{\pi}{2}v_{ik})}.
    \label{eq5:tf_v}
\end{equation}

Considering a uniformly distributed random number $\kappa \in (0,1)$, the positions of the agents in the binary space are updated according to
\begin{equation}
    x_{np,d}^{t+1} = 
    X(m,n)=
            \begin{cases}
                (x_{np,d}^{t})^{-1}, \ \quad \text{if } \kappa < S(v_{ik}^{t+1}) \\
                x_{np,d}^{t}, \quad \qquad \text{if } \kappa \geq S(v_{ik}^{t+1}).
            \end{cases}
        \label{eq5:binary_position}
\end{equation}

\section{Meta-Model Structure Selection (Meta-MSS): Building NARX for Regression}\label{met}

In this section, the use of a method based on meta-heuristic to select the NARX model structure is addressed. The BPSOGSA is implemented to search for the best model structure in a decision space formed by a predefined dictionary of regressors. The objective function of the optimization problem is based on the root mean squared error of the free run simulation output multiplied by a penalty factor that takes into account the complexity and the individual contribution of each regressor to build the final model.

\subsection{Encoding scheme}
The use of BPSOGSA for model structure selection is described. First, one should define the dimension of the test function. In this regard, the $n_y$,  $n_x$ and $\ell$ are set to generate all possibilities of regressors and a general matrix of regressors, $\Psi$, is built. The number of columns of $\Psi$ is assigned to the variable $noV$, and the number of agents, $N$, is defined. Then a binary $noV \times N$ matrix referred as $\mathcal{X}$, is randomly generated with the position of each agent in the search space. Each column of $\mathcal{X}$ represents a possible solution; in other words, a possible model structure to be evaluated at each iteration. Since each column of $\Psi$ corresponds a possible regressor, a value of $1$ in $\mathcal{X}$ indicates that, in its respective position, the column of $\Psi$ is included in the reduced matrix of regressors, while the value of $0$ indicates that the regressor column is ignored. 
\begin{exa}
\label{exa:encondig-scheme}
Consider a case where all possible regressors are defined based on $\ell ~=~1$ and $n_y ~=~ n_u ~=~ 2$. The $\Psi$ is defined by 
\begin{align}
[constant \ y(k-1) \ y(k-2) \ u(k-1) \ u(k-2)]
\end{align}

Because there are $5$ possible regressors, $noV = 5$. Assume $N = 5$, then $\mathcal{X}$ can be represented, for example, as \begin{equation}
    \mathcal{X} = 
    \begin{bmatrix}
0   &  1  &   0  &   0  &   0   \\
1   &  1  &   1  &   0  &   1   \\ 
0   &  0  &   1  &   1  &   0    \\
0   &  1  &   0  &   0  &   1    \\
1   &  0  &   1  &   1  &   0    
\end{bmatrix}
\end{equation}

The first column of $\mathcal{X}$ is transposed and used to generate a candidate solution: 
\begin{equation*}
    \mathcal{X} = 
    \begin{bmatrix}
constant & y(k-1) & y(k-2) & u(k-1) & u(k-2)   \\
1   &  1  &   1  &   0  &   1   
\end{bmatrix}
\end{equation*}

Hence, in this example, the first model to be tested is~$\alpha y(k-1) + \beta u(k-2)$, where~$\alpha$ and~$\beta$ are parameters estimated via Least Squares method. After that, the second column of~$X$ is tested and so on.
\label{exm}
\end{exa}

\subsection{Formulation of the objective function}
For each candidate model structure randomly defined, the linear-in-the-parameters system can be solved directly using the Least Squares algorithm. The variance of estimated parameters can be calculated as:
\begin{equation}
\label{eqmono:variance}
    \hat{\sigma}^2 = \hat{\sigma}_e^2V_{jj},
\end{equation}
\noindent where $\hat{\sigma}_e^2$ is the estimated noise variance calculated as
\begin{equation}
\label{eq:hat-sigma}
    \hat{\sigma}_e^2 = \frac{1}{N-m}\sum_{k=1}^{N}(y_k - \psi_{k-1}^\top\hat{\Theta})
\end{equation}
\noindent and $V_{jj}$ is the $jth$ diagonal element of $(\Psi^\top\Psi)^{-1}$.

The estimated standard error of the $j$th regression coefficient $\hat{\Theta}_j$ is the positive square root of the diagonal elements of $\hat{\sigma}^2$,
\begin{equation}
\label{eqmono:se}
    \mathrm{se}(\hat{\Theta}_j) = \sqrt{\hat{\sigma}^2_{jj}}. 
\end{equation}

A penalty test considers the standard error of the regression coefficients to determine the statistical relevance of each regressor. The $t$-test is used in this study to perform a \textit{hypothesis test} on the coefficients to check the significance of individual regressors in the multiple linear regression model. The hypothesis statements involve testing the \textit{null hypothesis} described as:
\begin{align*}
   \bm{H_0} & : \Theta_j = 0,
   \\
   \bm{H_a} & : \Theta_j \neq 0.
\end{align*}

In practice, one can compute a $t$-statistic as
\begin{equation}
\label{eqmono:t-value}
    T_0 = \frac{\hat{\Theta}_j}{\mathrm{se}(\hat{\Theta})},
\end{equation}
\noindent which measures the number of standard deviations that $\hat{\Theta}_j$ is away from $0$. More precisely, let
\begin{equation}
\label{eq:accept_region}
    -t_{\alpha/2, N-m} < T < t_{\alpha/2, N-m},
\end{equation}
\noindent where $t_{\alpha/2, N-m}$ is the $t$ value obtained considering $\alpha$ as the significance level and $N-m$ the degree of freedom. Then, If $T_0$ does not lie in the acceptance region of Eq.~\eqref{eq:accept_region}, the \textit{null hypothesis}, $\bm{H_0}: \Theta_j = 0$, is rejected and it is concluded that $\Theta_j$ is significant at $\alpha$. Otherwise, $\theta_j$ is not significantly different from zero, and the null hypothesis $\theta_j = 0$ cannot be rejected.

\subsubsection{Penalty value based on the Derivative of the Sigmoid Linear Unit function}

We proposed a penalty value based on the derivative of Eq.~\eqref{silu-shape} defined as:
\begin{equation}
\label{eq:dsigmoid}
    \dot{\varsigma}(x(\varrho)) = \varsigma(x)[1 + (a(x-c))(1 - \varsigma(x))]. 
\end{equation}

In this respect, the parameters of Eq.~\eqref{eq:dsigmoid} are defined as follows: $x$ has the dimension of $noV$; $c = noV/2$; and $a$ is defined by the number of regressors of the current test model divided by $c$. This approach results in a different curve for each model, considering the number of regressors of the current model. As the number of regressor increases, the slope of the sigmoid curve becomes steeper. The penalty value, $\varrho$, corresponds to the value in $y$ of the correspondent sigmoid curve regarding the number of regressor in $x$. It is imperative to point out that because the derivative of the sigmoid function return negative values, we normalize $\varsigma$ as
\begin{equation}
    \varrho = \varsigma - \mathrm{min}(\varsigma),
\end{equation}
\noindent so $\varrho \in \mathbb{R}^{+}$.

However, two different models can have the same number of regressors and present significantly different results. This situation can be explained based on the importance of each regressor in the composition of the model. In this respect, we use the $t$-student test to determine the statistical relevance of each regressor and introduce this information on the penalty function. In each case, the procedure returns the number of regressors that are not significant for the model, which we call $n_{\Theta, H_{0}}$. Then, the penalty value is chosen considering the model sizes as 
\begin{equation}
    \mathrm{model\_size} = n_{\Theta} + n_{\Theta, H_{0}}. 
\end{equation}

The objective function considers the relative root squared error of the model and $\varrho$ and is defined as
\begin{equation}
        \mathcal{F}= \frac{\sqrt{\sum\limits_{k=1}^{n}(y_k-\hat{y}_k)^2}}{\sqrt{\sum\limits_{k=1}^{n}(y_k-\bar{y})^2}} \times \varrho.
\label{eq:obj-function}
\end{equation}

With this approach, even if the tested models have the same number of regressors, the model which contain redundant regressors are penalized with a more substantial penalty value.

Finally, the Algorithm $6$ summarizes the method.

\begin{algorithm}[htpb]
\label{algo:mmss}
\DontPrintSemicolon
\SetAlgoLined
\KwResult{Model which has the best fitness value}
\KwIn{$\{(u_k), (y_k), k = 1, \ldots, N \}$, $\mathcal{M} = \{\psi_j, j = 1, \ldots, m\}$, $n_y$, $n_u$, $\ell$, max\_iteration, $noV$, np}

$\mathcal{P} \leftarrow$ Build initial population of random agents in the search space, $\mathcal{S}$ \;

$v \leftarrow$ set the agent's velocity equal zero at first iteration \; 

$\Psi \leftarrow$ Build the general matrix of regressors based on $n_y$, $n_u$ and $\ell$

\Repeat{max\_iterations is reached}{
\For{$i = 1:d$}{
$\mathrm{m_i} \leftarrow$ $\vec{x}_{\,np,i}$ \Comment*[r]{Extract the model encoding from population}

$\Psi_r \leftarrow$ $\Psi (\mathrm{m_i})$\Comment*[r]{Delete the $\Psi$ columns where $\mathrm{m_i} = 0$ \quad Ex.\ref{exa:encondig-scheme}}

$\hat{\Theta} \leftarrow$  $ (\Psi_r^\top\Psi_r)^{-1}\Psi_r^\top y$

$\hat{y} \leftarrow$ Free-run simulation of the model


$V \leftarrow (\Psi^\top\Psi)^{-1}$

$\hat{\sigma}_e^2 = \frac{1}{N-m}\sum_{k=1}^{N}(y_k - \psi_{k-1}^\top\hat{\Theta})$ \Comment*[r]{Eq.\ref{eq:hat-sigma}}

\For{$h = 1:\tau$}{
$\hat{\sigma}^2 \leftarrow \hat{\sigma}_e^2V_{h,h}$  \Comment*[r]{Eq.\ref{eqmono:variance}}

$\mathrm{se}(\hat{\Theta}_j) \leftarrow \sqrt{\hat{\sigma}^2_{h,h}}$ \Comment*[r]{Eq.\ref{eqmono:se}}

$T_0 \leftarrow \frac{\hat{\Theta}_j}{\mathrm{se}(\hat{\Theta})}$ \Comment*[r]{Eq.\ref{eqmono:t-value}}

$p \leftarrow$ regressors where $-t_{\alpha/2, N-m} < T_0 < t_{\alpha/2, N-m}$ \Comment*[r]{Eq.\ref{eq:accept_region}}
}
Remove the $p$ regressors from $\Psi_r$\;

Check for empty model\;

\If{\text{Model is empty}}{
Generate a new population\; 

Repeat the steps from line $6$ to $18$

}

$n_1 \leftarrow$ size($p$) \Comment*[r]{Number of redundant terms}

$\hat{\Theta} \leftarrow$  $ (\Psi_r^\top\Psi_r)^{-1}\Psi_r^\top y$ \Comment*[r]{Re-estimation}

$\mathcal{F}_i \leftarrow \frac{\sqrt{\sum\limits_{k=1}^{n}(y_k-\hat{y}_k)^2}}{\sqrt{\sum\limits_{k=1}^{n}(y_k-\bar{y})^2}} \times \varrho$ \Comment*[r]{Eq.\ref{eq:obj-function}}

$\mathcal{P}^n_{i} \leftarrow $ Encoded $\Psi_r$\;

Evaluate the fitness for each agent, $\mathcal{F}_i(t)$
}

$\mathcal{P} \leftarrow \mathcal{P}^n$ \Comment*[r]{Update the population}

\ForEach{$\vec{x}_{\,np,d} \in \mathcal{P}$}{
Calculate the acceleration of each agent \Comment*[r]{Eq.\ref{eq5:accelaration}}
Adapt the $\mathrm{c}_j'$ coefficients \Comment*[r]{Eq.\ref{eq5:adapt_c}}
Update the velocity of the agents \Comment*[r]{Eq.\ref{eq5:vel_psogsa}}
Update the position of the agents \Comment*[r]{Eq.\ref{eq5:vel_pos_gsa}}
}
}
\caption{Meta-structure selection (Meta-MSS) algorithm}
\end{algorithm}

\subsection{Case Studies: Simulation Results}
In this section, six simulation examples are considered to illustrate the effectiveness of the Meta-MSS algorithm. An analysis of the algorithm performance has been carried out considering different tuning parameters. The selected systems are generally used as a benchmark for model structures algorithms and were taken from~\cite{WB2008,FPP2015,BAK2013,PS2003,GGBW2015,BSP2010,ABB2010}. Finally, a comparative analysis with respect to the Randomized Model Structure Selection (RaMSS)~\cite{FPP2015}, the FROLS~\cite{Bil2013}, and the Reversible-jump Markov chain Monte Carlo (RJMCMC)~\cite{BAK2013} algorithms has been accomplished to check out the goodness of the proposed method. 

The simulation models are described as:
{\small
\begin{align}
\label{simulated_systems}
    & S_1: \quad y_k = -1.7y_{k-1} - 0.8y_{k-2} + x_{k-1} + 0.81x_{k-2} + e_k, \\
    & \qquad \quad \text{with } x_k \sim \mathcal{U}(-2, 2) \text{ and } e_k \sim \mathcal{N}(0, 0.01^2); \nonumber \\
    & S_2: \quad y_k = 0.8y_{k-1} + 0.4x_{k-1} + 0.4x_{k-1}^2 + 0.4x_{k-1}^3 + e_k, \\
    & \qquad \quad \text{with } x_k \sim \mathcal{N}(0, 0.3^2) \text{ and } e_k \sim \mathcal{N}(0, 0.01^2). \nonumber \\
    & S_3: \quad y_k = 0.2y_{k-1}^3 + 0.7y_{k-1}x_{k-1} + 0.6x_{k-2}^2 \nonumber \\
    &- 0.7y_{k-2}x_{k-2}^2 -0.5y_{k-2}+ e_k, \\
    & \qquad \quad \text{with } x_k \sim \mathcal{U}(-1, 1) \text{ and } e_k \sim \mathcal{N}(0, 0.01^2). \nonumber \\
    & S_4: \quad y_k = 0.7y_{k-1}x_{k-1} - 0.5y_{k-2} + 0.6x_{k-2}^2 \nonumber \\ 
    &- 0.7y_{k-2}x_{k-2}^2 + e_k, \\
    & \qquad \quad \text{with } x_k \sim \mathcal{U}(-1, 1) \text{ and } e_k \sim \mathcal{N}(0, 0.04^2). \nonumber \\
    & S_5: \quad y_k = 0.7y_{k-1}x_{k-1} - 0.5y_{k-2} + 0.6x_{k-2}^2 \nonumber \\
    &- 0.7y_{k-2}x_{k-2}^2 + 0.2e_{k-1} \nonumber \\ 
    & \qquad \quad - 0.3x_{k-1}e_{k-2} + e_k,\\
    & \qquad \quad \text{with } x_k \sim \mathcal{U}(-1, 1) \text{ and } e_k \sim \mathcal{N}(0, 0.02^2); \nonumber \\
    & S_6: \quad y_k = 0.75y_{k-2} + 0.25x_{k-2} - 0.2y_{k-2}x_{k-2} + e_k \nonumber \\ 
    & \qquad \quad \text{with } x_k \sim \mathcal{N}(0, 0.25^2) \text{ and } e_k \sim \mathcal{N}(0, 0.02^2); \nonumber
\end{align}
}%
\noindent where $\mathcal{U}(a, b)$ are samples evenly distributed over~$[a, b]$, and $\mathcal{N}(\eta, \sigma^2)$ are samples with a Gaussian distribution with mean $\eta$ and standard deviation $\sigma$. All realizations of the systems are composed of a total of $500$ input-output data samples. Also, the same random seed is used to reproducibility purpose.

All tests have been performed in Matlab\textregistered~$2018$a environment, on a Dell Inspiron $5448$ Core i$5-5200$U CPU $2.20$GHz with $12$GB of RAM.

Following the aforementioned studies, the maximum lags for the input and output are chosen to be, respectively,~$n_u=n_y=4$ and the nonlinear degree is~$\ell = 3$. The parameters related to the BPSOGSA are detailed on Table~\eqref{tab:pbsogsa-parameters}. 

\begin{table}[!htb]
\centering
\caption{Parameters used in Meta-MSS}
\label{tab:pbsogsa-parameters}
\resizebox{0.48\textwidth}{!}{%
\begin{tabular}{ccccccccc}
\hline
Parameters & $n_u$ & $n_y$ & $\ell$ & p-value & max\_iter & n\_agents & $\alpha$ & $G_0$  \\ \hline
Values     & $4$  & $4$  & $3$ & $0.05$  & $30$      & $10$       & $23$    & $100$ \\ \hline
\end{tabular}%
}
\end{table}

$300$ runs of the Meta-MSS algorithm have been executed for each model, aiming to compare some statistics about the algorithm performance. The elapsed time, the time required to obtain the final model, and correctness, the percentage of exact model selections, are analyzed.

The results in Table~\ref{tab:results-meta-mss} are obtained with the parameters configured accordingly to Table~\eqref{tab:pbsogsa-parameters}.
\begin{table}[!htbp]
\centering
\caption{Overall performance of the Meta-MSS}
\label{tab:results-meta-mss}
\resizebox{0.48\textwidth}{!}{%
\begin{tabular}{ccccccc}
\hline
              & $S_1$  & $S_2$  & $S_3$  & $S_4$  & $S_5$ & $S_6$ \\ \hline
Correct model & $100\%$ & $100\%$ & $100\%$ & $100\%$ & $100\%$ & $100\%$ \\
Elapsed time (mean) & $5.16$s   & $3.90$s   & $3.40$s   & $2.37$s   & $1.40$s & $3.80$s  \\ \hline
\end{tabular}%
}
\end{table}

Table~\ref{tab:results-meta-mss} shows that all the model terms are correctly selected using the Meta-MSS. It is worth to notice that even the model $S_5$, which have an autoregressive noise, was correctly selected using the proposed algorithm. This result resides in the evaluation of all regressors individually, and the ones considered redundant are removed from the model. 

\begin{figure*}[!htb]
    \centering
    \begin{subfigure}[b]{0.3\textwidth}
        \centering
        \includegraphics[width=\textwidth, height = 3.2cm]{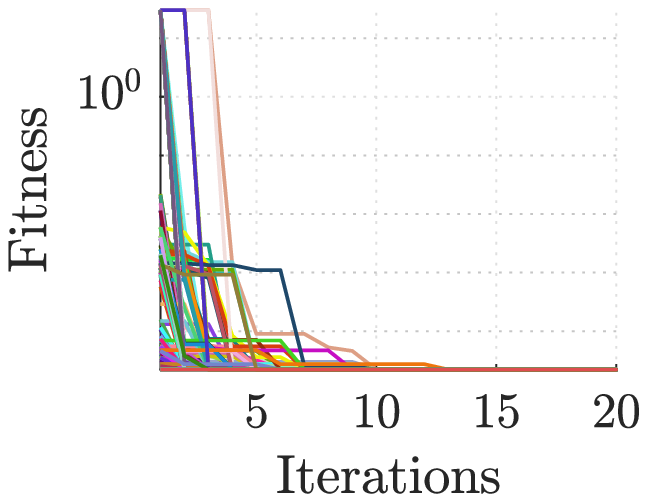}
        \caption{System $S_1$.}
        \label{fig:cs1}
    \end{subfigure}
    \hfill
    \begin{subfigure}[b]{0.3\textwidth}
        \centering
        \includegraphics[width=\textwidth, height = 3.2cm]{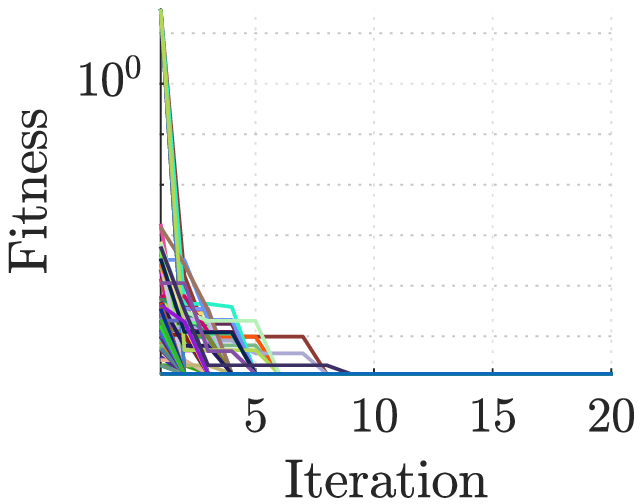}
        \caption{System $S_2$.}
        \label{fig:cs2}
    \end{subfigure}
     \hfill
    \begin{subfigure}[b]{0.3\textwidth}
        \centering
        \includegraphics[width=\textwidth, height = 3.2cm]{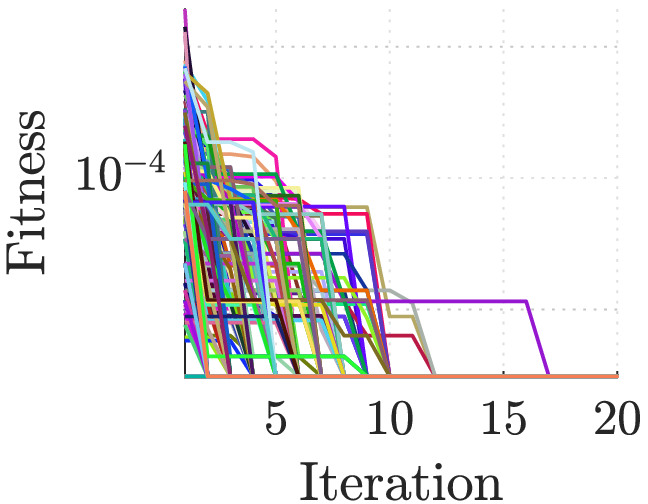}
        \caption{System $S_3$.}
        \label{cs3}
    \end{subfigure}
     \hfill
    \begin{subfigure}[b]{0.3\textwidth}
        \centering
        \includegraphics[width=\textwidth, height = 3.2cm]{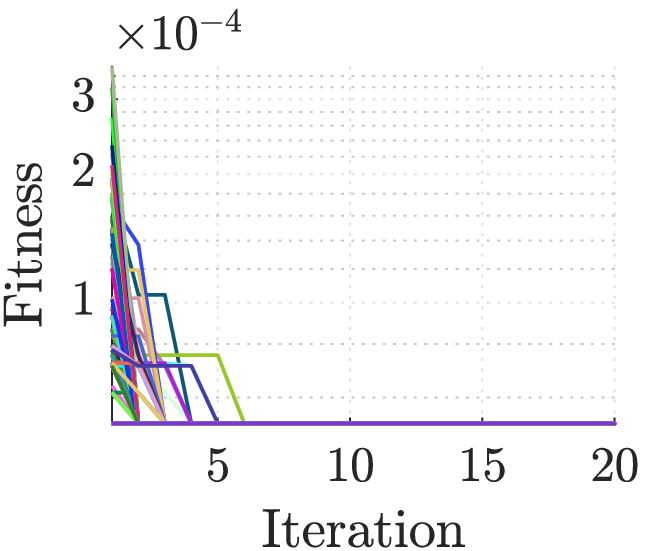}
        \caption{System $S_4$.}
        \label{fig:cs4}
    \end{subfigure}
     \hfill
    \begin{subfigure}[b]{0.3\textwidth}
        \centering
        \includegraphics[width=\textwidth, height = 3.2cm]{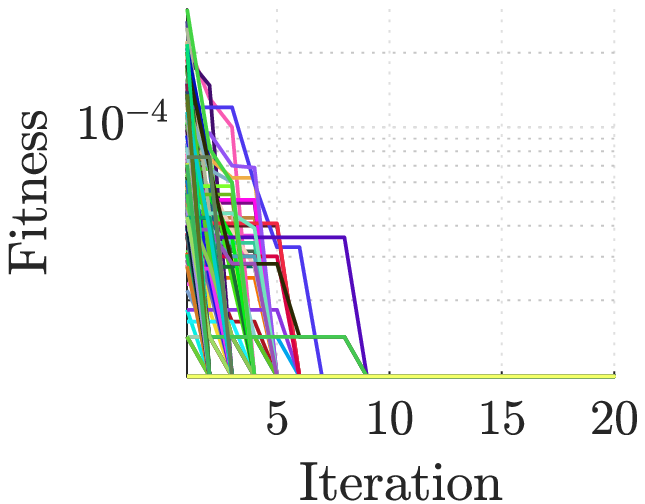}
        \caption{System $S_5$.}
        \label{fig:cs5}
    \end{subfigure}
     \hfill
    \begin{subfigure}[b]{0.3\textwidth}
        \centering
        \includegraphics[width=\textwidth, height = 3.2cm]{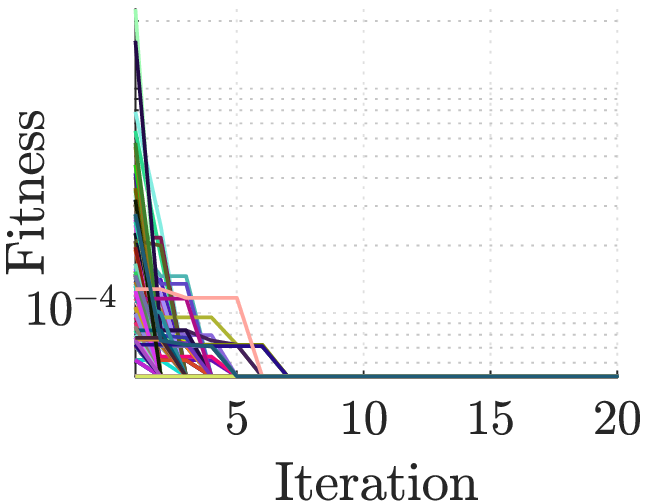}
        \caption{System $S_6$.}
        \label{fig:cs6}
    \end{subfigure}
    \caption[Comparison of the convergence curves]{The convergence of each execution of Meta-MSS algorithm.}
    \label{fig:convergence-s1-s5-30}
\end{figure*}

Figure~\ref{fig:convergence-s1-s5-30} present the convergence of each execution of Meta-MSS. It is noticeable that the majority of executions converges to the correct model structures with $10$ or fewer iterations. The reason for this relies on the maximum number of iterations and the number of search agents. The first one is related to the acceleration coefficient, which boosts the exploration phase of the algorithm, while the latter increases the number of candidate models to be evaluated. Intuitively, one can see that both parameters influence the elapsed time and, more importantly, the model structure selected to compose the final model. Consequently, an inappropriate choice of one of them may results in sub/over-parameterized models, since the algorithm can converge to a local optimum. The next subsection presents an analysis of the max\_iter and n\_agents influence in the algorithm performance.

\subsection{Meta-MSS vs RaMSS vs C-RaMSS}

The systems $S_1$, $S_2$, $S_3$, $S_4$ and $S_6$ has been used as benchmark by~\cite{BFPP2017}, so we can compare directly our results with those reported by the author in his thesis. All techniques used $n_y=n_u=4$ and $\ell = 3$. The RaMSS and the RaMSS with Conditional Linear Family (C-RaMSS) used the following configuration for the tuning parameters: $K=1$, $\alpha = 0.997$, $NP = 200$ and $v=0.1$. The Meta-Structure Selection Algorithm was tuned according to Table~\ref{tab:pbsogsa-parameters}.
\begin{table*}[!htb]
\centering
\caption{Comparative analysis between Meta-MSS, RaMSS, and C-RaMSS}
\label{tab:meta-ramss-cramss}
\begin{tabular}{lllllll}
\hline
                                                      & \multicolumn{1}{c}{}                    & \multicolumn{1}{c}{$S_1$}   & \multicolumn{1}{c}{$S_2$}   & \multicolumn{1}{c}{$S_3$}   & \multicolumn{1}{c}{$S_4$}   & \multicolumn{1}{c}{$S_6$}   \\ \hline
\multicolumn{1}{l|}{\multirow{2}{*}{Meta-MSS}}        & \multicolumn{1}{c}{Correct model}       & \multicolumn{1}{c}{$100\%$} & \multicolumn{1}{c}{$100\%$} & \multicolumn{1}{c}{$100\%$} & \multicolumn{1}{c}{$100\%$} & \multicolumn{1}{c}{$100\%$} \\
\multicolumn{1}{l|}{}                                 & \multicolumn{1}{c}{Elapsed time (mean)} & \multicolumn{1}{c}{$5.16$s} & \multicolumn{1}{c}{$3.90$s} & \multicolumn{1}{c}{$3.40$s} & \multicolumn{1}{c}{$2.37$s} & \multicolumn{1}{c}{$3.80$s} \\ \hline
\multicolumn{1}{l|}{\multirow{2}{*}{RaMSS- $NP=100$}} & Correct model                           & $90.33\%$                   & $100\%$                     & $100\%$                     & $100\%$                     & $66\%$                      \\
\multicolumn{1}{l|}{}                                 & Elapsed time (mean)                     & $3.27$s                     & $1.24$s                     & $2.59$s                     & $1.67$s                     & $6.66$s                     \\ \hline
\multicolumn{1}{l|}{\multirow{2}{*}{RaMSS- $NP=200$}} & Correct model                           & $78.33\%$                   & $100\%$                     & $100\%$                     & $100\%$                     & $82\%$                      \\
\multicolumn{1}{l|}{}                                 & Elapsed time (mean)                     & $6.25$s                     & $2.07$s                     & $4.42$s                     & $2.77$s                     & $9.16$s                     \\ \hline
\multicolumn{1}{l|}{\multirow{2}{*}{C-RaMSS}}         & Correct model                           & $93.33\%$                   & $100\%$                     & $100\%$                     & $100\%$                     & $100\%$                     \\
\multicolumn{1}{l|}{}                                 & Elapsed time (mean)                     & $18$s                       & $10.50$s                    & $16.96$s                    & $10.56$s                    & $48.52$s                    \\ \hline
\end{tabular}%
\end{table*}

In terms of correctness, the Meta-MSS outperforms (or at least equals) the RaMSS and C-RaMSS for all analyzed systems as shown in Table~\ref{tab:meta-ramss-cramss}. Regarding $S_6$, the correctness rate increased by $18\%$ when compared with RaMSS and the elapsed time required for C-RaMSS obtain $100\%$ of correctness is $1276.84\%$ higher than the Meta-MSS. Furthermore, the Meta-MSS is notably more computationally efficient than C-RaMSS and similar to RaMSS.

\subsection{Meta-MSS vs FROLS}
The FROLS algorithm has been tested on all the systems and the results are detailed in Table~\ref{tab:meta-frols}. It can be seen that only the model terms selected for $S_2$ and $S_6$ are correct using FROLS. The FROLS fails to select two out of four regressors for $S_1$. Regarding $S_3$, the term $y_{k-1}$ is included in the model instead of $y_{k-1}^3$. Similarly, the term $y_{k-4}$ is wrongly added in model $S_4$ instead of $y_{k-2}$. Finally, an incorrect model structure is returned for $S_5$ as well with the addition of the spurious term $y_{k-4}$. 

\begin{table}[!htb]
\centering
\caption{Comparative analysis - Meta-MSS vs FROLS}
\caption{C}
\label{tab:meta-frols}
\resizebox{0.49\textwidth}{!}{%
\begin{tabular}{ccccc}
\hline
                       & \multicolumn{2}{c}{Meta-MSS} & \multicolumn{2}{c}{FROLS}    \\ \hline
                       & Regressor          & Correct & Regressor          & Correct \\ \hline
\multirow{4}{*}{$S_1$} & $y_{k-1}$          & yes     & $y_{k-1}$          & yes     \\
                       & $y_{k-2}$          & yes     & $y_{k-4}$          & no      \\
                       & $x_{k-1}$          & yes     & $x_{k-1}$          & yes     \\
                       & $x_{k-2}$          & yes     & $x_{k-4}$          & no     \\ \hline
\multirow{4}{*}{$S_2$} & $y_{k-1}$          & yes     & $y_{k-1}$          & yes     \\
                       & $x_{k-1}$          & yes     & $x_{k-1}$          & yes     \\
                       & $x_{k-1}^2$        & yes     & $x_{k-1}^2$        & yes     \\
                       & $x_{k-1}^3$        & yes     & $x_{k-1}^3$        & yes     \\ \hline
\multirow{5}{*}{$S_3$} & $y_{k-1}^3$        & yes     & $y_{k-1}  $        & no     \\
                       & $y_{k-1}x_{k-1}$   & yes     & $y_{k-1}x_{k-1}$   & yes     \\
                       & $x_{k-2}^2$        & yes     & $x_{k-2}^2$        & yes     \\
                       & $y_{k-2}x_{k-2}^2$ & yes     & $y_{k-2}x_{k-2}^2$ & yes     \\
                       & $y_{k-2}$          & yes     & $y_{k-2}$          & yes     \\ \hline
\multirow{4}{*}{$S_4$} & $y_{k-1}x_{k-1}$   & yes     & $y_{k-1}x_{k-1}$   & yes     \\
                       & $y_{k-2}$          & yes     & $y_{k-4}$          & no     \\
                       & $x_{k-2}^2$        & yes     & $x_{k-2}^2$        & yes     \\
                       & $y_{k-2}x_{k-2}^2$ & yes     & $y_{k-2}x_{k-2}^2$ & yes     \\ \hline
\multirow{4}{*}{$S_5$} & $y_{k-1}x_{k-1}$   & yes     & $y_{k-1}x_{k-1}$   & yes     \\
					   & $y_{k-2}$          & yes     & $y_{k-4}$          & no     \\
				       & $x_{k-2}^2$        & yes     & $x_{k-2}^2$        & yes     \\
					   & $y_{k-2}x_{k-2}^2$ & yes     & $y_{k-2}x_{k-2}^2$ & yes     \\ \hline
\multirow{3}{*}{$S_6$} & $y_{k-2}$          & yes     & $y_{k-2}$          & yes     \\
                       & $x_{k-1}$          & yes     & $x_{k-1}$          & yes     \\
                       & $y_{k-2}x_{k-2}$   & yes     & $y_{k-2}x_{k-1}$   & yes     \\ \hline
\end{tabular}%
}
\end{table}

\subsection{Meta-MSS vs RJMCMC}
The $S_4$ is taken from \cite{BAK2013}. Again the maximum lag for the input and output are~$n_y = nu = 4$ and the nonlinear degree is~$\ell = 3$. In their work, the authors executed the algorithm $10$ times on the same input-output data. The RJMCMC was able to select the true model structure $7$ times out of the $10$ runs. On the other hand, the Meta-MSS can get the true model in all runs of the algorithm. The results are summarized in Table~\ref{tab:meta-rjmcmc}. Besides, there are main drawbacks related to RJMCMC method which are overcome by the Meta-MSS: the former is computationally expensive and required an execution with $30,000$ iterations. Furthermore, it assumes different probability distributions which are chosen to ease the computations for the parameters involved in the procedure.

\begin{table*}[!htb]
	\centering
	\caption{Comparative analysis - Meta-MSS vs RJMCMC}
	\label{tab:meta-rjmcmc}
		\begin{tabular}{cccccccc}
			\hline
			& \multicolumn{2}{c}{Meta-MSS} & \multicolumn{5}{c}{RJMCMC}                                                                   \\ \hline
			& Model              & Correct & Model 1 ($7\times$) & Model 2            & Model 3            & Model 4            & Correct \\ \hline
			\multirow{5}{*}{$S_4$} & $y_{k-1}x_{k-1}$   & yes     & $y_{k-1}x_{k-1}$    & $y_{k-1}x_{k-1}$   & $y_{k-1}x_{k-1}$   & $y_{k-1}x_{k-1}$   & yes     \\
			& $y_{k-2}$          & yes     & $y_{k-2}$           & $y_{k-2}$          & $y_{k-2}$          & $y_{k-2}$          & yes     \\
			& $x_{k-2}^2$        & yes     & $x_{k-2}^2$         & $x_{k-2}^2$        & $x_{k-2}^2$        & $x_{k-2}^2$        & yes     \\
			& $y_{k-2}x_{k-2}^2$ & yes     & $y_{k-2}x_{k-2}^2$  & $y_{k-2}x_{k-2}^2$ & $y_{k-2}x_{k-2}^2$ & $y_{k-2}x_{k-2}^2$ & yes     \\
			& -                  & -       & -                   & $y_{k-3}x_{k-3}$   & $x_{k-4}^2$        & $x_{k-1}x_{k-3}^2$ & no      \\ \hline
		\end{tabular}%
\end{table*}

\subsection{Full-scale F-16 aircraft}
The F-16 Ground Vibration Test has been used as a benchmark for system identification. The case exhibits a clearance and friction nonlinearities at the mounting interface of the payloads. The empirical data were acquired on a full-scale F-16 aircraft on a Siemens LMS Ground Vibration Testing Master Class as well as a detailed formulation of the identification problem is available at Nonlinear System Identification Benchmarks \footnote{Available at http://www.nonlinearbenchmark.org/}.

Several datasets are available concerning different input signals and frequencies. This work considers the data recorded under multisine excitations with a full frequency grid from $2$ to $15 Hz$. According to~\cite{NS2017}, at each force level, $9$ periods were acquired considering a single realization of the input signal. There are $8192$ samples per period. Note that transients are present in the first period of measurement.

This case study represents a significant challenge because it involves nonparametric analysis of the data, linearized modeling, and damping ratios versus the excitation level and nonlinear modeling around a single mode. Also, the order of the system is reasonably high. In the $2 - 15 Hz$ band, the F-16 possesses about $10$ resonance modes. 
\begin{table}[!ht]
	\centering
	\caption{Parameters used in Meta-Structure Selection Algorithm for the F-16 benchmark}
	\label{tab:pbsogsa-parameters-f16}
	\resizebox{0.49\textwidth}{!}{%
	\begin{tabular}{cccccccccc}
		\hline
		Parameters & $n_u$ & $n_{u2}$ & $n_y$ & $\ell$ & p-value & max\_iter & n\_agents & $\alpha$ & $G_0$  \\ \hline
		Values     & $20$  & $20$   & $20$  & $1$ & $0.05$  & $30$      & $15$       & $23$    & $100$ \\ \hline
	\end{tabular}%
	}
\end{table}

The Meta-MSS algorithm and the FROLS are used to select models to represent the dynamics of the F-16 aircraft described above. In the first approach, the maximum nonlinearity degree and the lag of inputs and output were set to $2$ and $10$, respectively. In this case, the Meta-MSS select a model with $15$ terms, but the model selected through FROLS diverged. Thus, we set the maximum lag to $2$. The Meta-MSS has chosen $3$ regressors to form the model, while the FROLS failed again to build an adequate model. Finally, the maximum lag was set to $20$ and the maximum nonlinearity degree was defined to be $1$. For the latter case, the Meta-MSS parameters are defined as listed in Table~\ref{tab:pbsogsa-parameters-f16}. The same input and output lags are considered on FROLS approach. Table~\ref{tab:meta-frols-f16} details the results considering the second acceleration signals as output. For this case, following the recommendation in~\cite{NS2017}, the models are evaluated using the metric $e_{rms_t}$, which is defined as:
\begin{equation}
e_{rms_t} = \sqrt{\frac{1}{N}\sum\limits_{k=1}^{N}(y_k - \hat{y}_k)^2}.
\end{equation}

\begin{table*}[!thb]
	\centering
	\caption{Identified NARX model using Algorithm 6 and FROLS.}
	\label{tab:meta-frols-f16}
	\resizebox{\textwidth}{!}{%
		\begin{tabular}{ccccccccc}
			\hline
			& \multicolumn{2}{c}{Meta-MSS}       & \multicolumn{2}{c}{Meta-MSS | $\ell = n_{x_1} = n_{x_2} =  2$}      & \multicolumn{2}{c}{Meta-MSS | $\ell = 2$, $ n_{x_1} = n_{x_2} =  10$}       & \multicolumn{2}{c}{FROLS}          \\ \hline
			& Model term             & Parameter & Model term            & Parameter & Model term            & Parameter  & Model term             & Parameter \\ \hline
			& $y_{k-1}$              & $0.7166$  & $y_{k-2}$             & $0.6481$  & $y_{k-1}$             & $1.3442$   & $y_{k-1}$              & $1.7829$  \\
			& $y_{k-5}$              & $0.2389$  & $x_{\indices{_1}k-1}$ & $1.5361$  & $y_{k-2}$             & $-0.8141$  & $y_{k-2}$              & $-1.8167$ \\
			& $y_{k-8}$              & $-0.0716$ & $x_{\indices{_2}k-2}$ & $1.3857$  & $y_{k-4}$             & $0.3592$   & $y_{k-3}$              & $1.3812$  \\
			& $y_{k-13}$             & -$0.0867$ &                       &           & $x_{\indices{_1}k-6}$ & $14.8635$  & $y_{k-6}$              & $1.5213$  \\
			& $x_{\indices{_1}k-2}$  & $1.5992$  &                       &           & $x_{\indices{_1}k-7}$ & $-14.7748$ & $y_{k-9}$              & $0.3625$  \\
			& $x_{\indices{_1}k-13}$ & $-1.1414$ &                       &           & $x_{\indices{_2}k-1}$ & $-3.2129$  & $x_{\indices{_2}k-7}$  & $-2.4253$ \\
			& $x_{\indices{_2}k-4}$  & $2.2248$  &                       &           & $x_{\indices{_2}k-3}$ & $7.1903$   & $x_{\indices{_1}k-1}$  & $1.8534$  \\
			& $x_{\indices{_2}k-8}$  & $-0.8383$ &                       &           & $x_{\indices{_2}k-8}$ & $-4.0374$  & $x_{\indices{_2}k-3}$  & $1.9866$  \\
			& $x_{\indices{_2}k-13}$ & $-1.1189$ &                       &           &                       &            & $x_{\indices{_1}k-8}$  & $-1.5305$ \\
			&                        &           &                       &           &                       &            & $x_{\indices{_2}k-1}$  & $0.6547$  \\
			&                        &           &                       &           &                       &            & $y_{k-7}$              & $1.2767$  \\
			&                        &           &                       &           &                       &            & $y_{k-5}$              & $1.3378$  \\
			&                        &           &                       &           &                       &            & $y_{k-10}$             & $-0.3234$ \\
			&                        &           &                       &           &                       &            & $y_{k-4}$              & $-1.0199$ \\
			&                        &           &                       &           &                       &            & $y_{k-8}$              & $-0.7116$ \\
			&                        &           &                       &           &                       &            & $y_{k-12}$             & $-0.2222$ \\
			&                        &           &                       &           &                       &            & $y_{k-11}$             & $0.3761$  \\
			&                        &           &                       &           &                       &            & $x_{\indices{_1}k-20}$ & $0.0245$  \\ \hline
			$e_{rms_t}$  & \multicolumn{2}{c}{$0.0862$}       & \multicolumn{2}{c}{$0.1268$}      & \multicolumn{2}{c}{$0.0982$}       & \multicolumn{2}{c}{$0.0876$}       \\ \hline
			Elapsed time & \multicolumn{2}{c}{$27.92s$}       & \multicolumn{2}{c}{$16.78$}       & \multicolumn{2}{c}{$207.13$}       & \multicolumn{2}{c}{$18.01s$}       \\ \hline
		\end{tabular}%
	}
\end{table*}
As highlighted in Table~\ref{tab:meta-frols-f16}, the Meta-MSS algorithm returns a model with $9$ regressors and a better performance than the model with $18$ terms built using FROLS. The step-by-step procedure used by FROLS results in the selection of the first $12$ output terms, while only $4$ output regressors are selected using Meta-MSS. From Table~\ref{tab:meta-frols-f16}, one can see that the Meta-MSS algorithm have an affordable computational cost, since the time to select the model is very acceptable, even when comparing with FROLS, which is known to be one of the most efficient methods for structure selection.

\begin{figure}[!ht]
	\centering
	\begin{subfigure}[b]{0.49\textwidth}
		\centering
		\includegraphics[width=\textwidth]{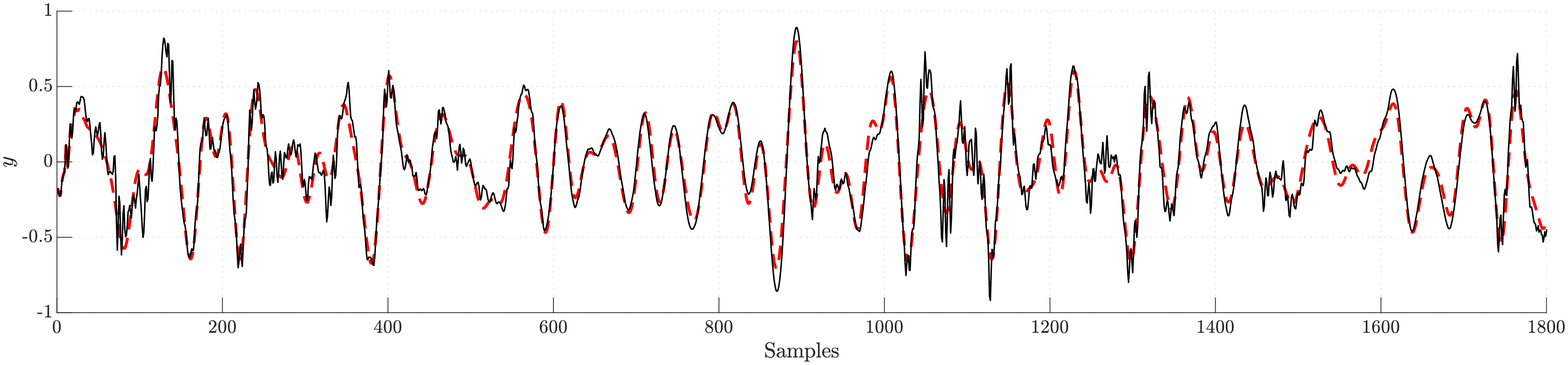}
		\caption{Meta-MSS: $\ell=1$, $ny = n_{x_1} = n_{x_2} = 20$.}
		\label{fig:f16meta}
	\end{subfigure}
	\hfill
	\begin{subfigure}[b]{0.49\textwidth}
		\centering
		\includegraphics[width=\textwidth]{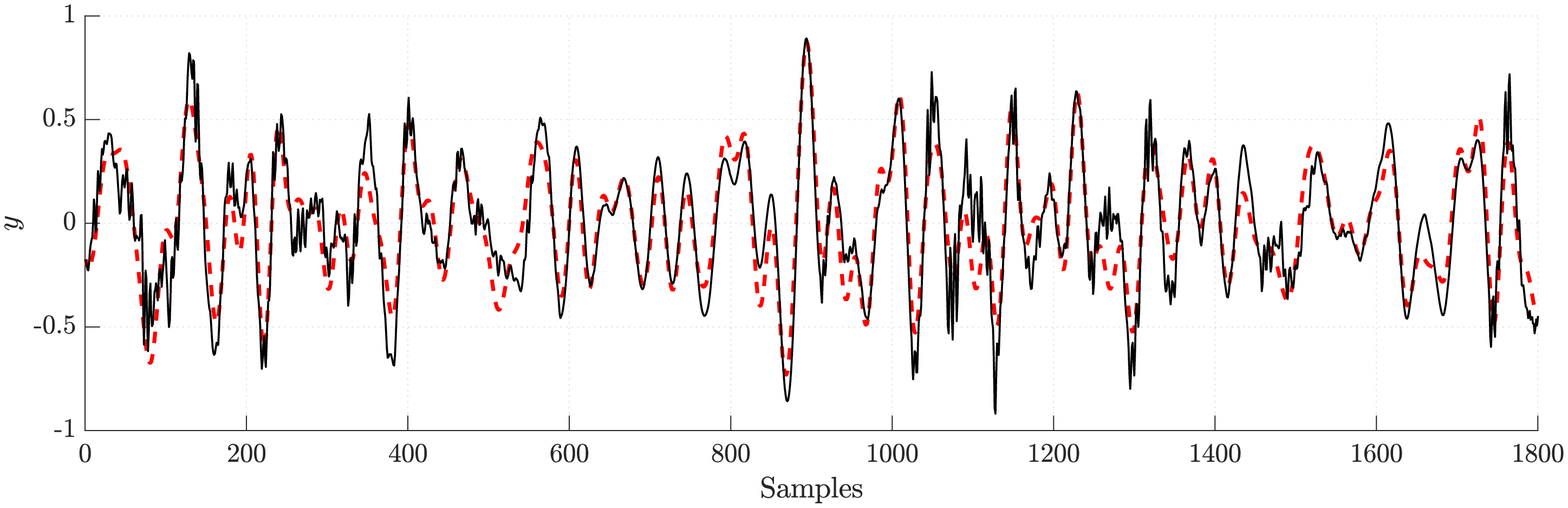}
		\caption{Meta-MSS: $\ell=2$, $n_y = n_{x_1} = n_{x_2} = 2$.}
		\label{fig:f16meta2}
	\end{subfigure}
	\hfill
	\begin{subfigure}[b]{0.49\textwidth}
		\centering
		\includegraphics[width=\textwidth]{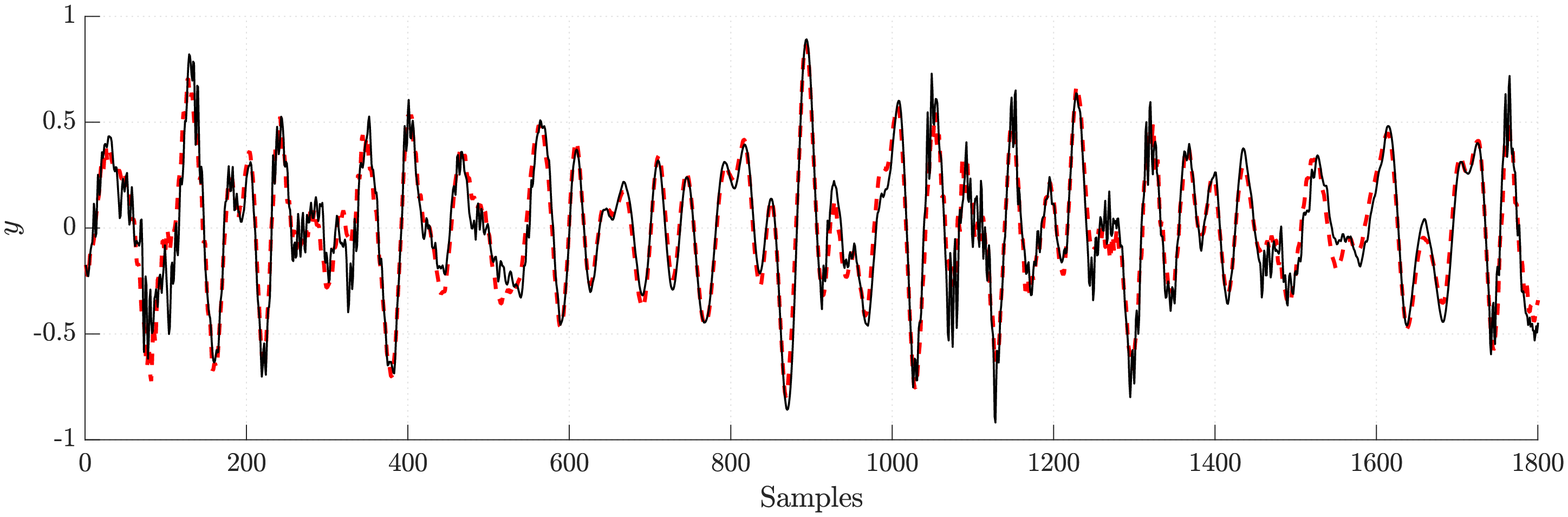}
		\caption{Meta-MSS: $\ell=2$, $n_y = n_{x_1} = n_{x_2} = 10$.}
		\label{fig:f16meta3}
	\end{subfigure}
	\hfill
	\begin{subfigure}[b]{0.49\textwidth}
		\centering
		\includegraphics[width=\textwidth]{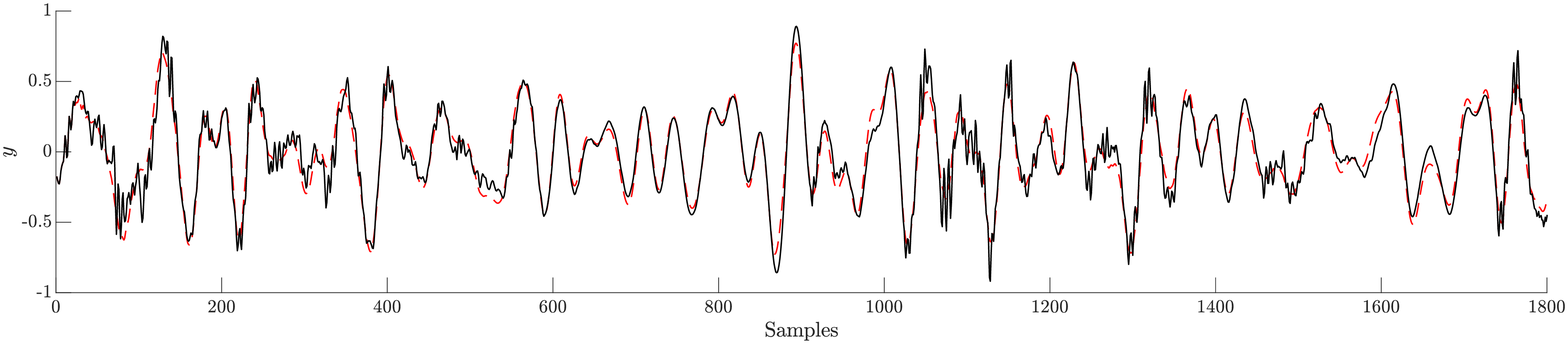}
		\caption{FROLS: $\ell=1$, $ny = n_{x_1} = n_{x_2} = 20$.}
		\label{fig:f16frols}
	\end{subfigure}
	\caption[Case Study: F-16 aircraft benchmark]{Models obtained using the Meta-MSS and the FROLS algorithm. The  FROLS was only capable to return a stable model when setting $\ell = 1$. The Meta-MSS, otherwise, returned satisfactory models in all cases.}
	\label{fig:meta-frols-f16}
\end{figure}

Further, it is interesting to note that the Meta-MSS returned a linear model even when the tests were performed using the maximum nonlinearity degree $\ell = 2$. This result demonstrates the excellent performance of the method since the classical one was not able to reach a satisfactory result. Figure~\ref{fig:meta-frols-f16} depicts the free run simulation of each model.

\section{Meta-Model Structure Selection (Meta-MSS): Building NARX for Classification}\label{met3}

Because of many real-life problems associate continuous and discrete variables, classification has been one of the most widely studied techniques for decision-making tasks in engineering, health science, business and many more. Many methods and algorithms have been developed to data classification,  which cover logistic regression~\cite{HLS2013}, random forest~\cite{Bre2001}, support vector machines~\cite{CS2000}, k-nearest neighbors~\cite{KJ2013} and logistic-NARX model for binary classification~\cite{SWB2019}. The former three algorithms are widely used, but the interpretation of such models is a hard task. Regarding logistic-NARX, besides the computational efficiency and transparency, it allows the inclusion of lagged terms straightforwardly while other techniques include lagged terms explicitly.

Following the logistic-NARX approach, this section adapts the Meta-MSS algorithm to develop NARX models focusing on the prediction of systems with binary responses that depend on continuous predictors. The primary motivation comes from the fact that the logistic-NARX approach inherits not only the goodness of the FROLS but all of its drawbacks related to being stocked in locally optimal solutions. A direct comparison with the methods above is performed using the identification and evaluation of two simulated models, and an empirical system.

\subsection{Logist NARX modeling approach using Meta-MSSc algorithm}

In~\cite{SWB2019}, the logistic-NARX is based on the FROLS algorithm to select the terms to compose the following probability model
\begin{equation}
p_k = \frac{1}{1+e^{-\psi^\top_{k-1}\hat{\Theta}}}.
\end{equation}

The biserial coefficient is used to measure the relationship between a continuous variable and a dichotomous variable according to~\cite{Pal2013}:
\begin{equation}
r(\bm{x},\bm{y}) = \frac{\overline{X}_1 - \overline{X}_0}{\sigma_X}\sqrt{\frac{n_1n_0}{N^2}},
\end{equation}
\noindent where $X_0$ is the mean value on the continuous variable $X$ for all the observations that belong to class $0$, $X_1$ is the mean value of variable $X$ for all the observations that belong to class $1$, $\sigma_X$ is the standard deviation of variable $X$, $n_0$ is the number of observations that belong to class $0$, $n_1$ is the number of observations that belong to class $1$, and $N$ is the total number of data points. Even though it is based on FROLS, the logistic-NARX approach requires the user to set a maximum number of regressors to form the final model, which is not required when using Meta-MSS algorithm for binary classification.

The objective function of the Meta-MSS is adapted to use the biserial correlation to measure the association between the variables instead of the RMSE.
For the continuous regression problem, the parameters are estimated using the LS method, which minimizes the sum of squared errors of the model output. Because we are dealing with categorical response variables, this approach is not capable of producing minimum variance unbiased estimators, so the parameters are estimated via a Stochastic Gradient Descent (SGD)~\cite{Bot2012}:

Apart from those changes, the main aspects of the standard Meta-MSS algorithm are held, such the regressor significance evaluation and all aspects of exploration and exploitation of the search space. Because the parameters are now estimated using SGD, the method becomes more computationally demanding, and this can slow down the method, especially when concerning with large models. 

\subsection{Electroencephalography Eye State Identification}
This dataset was built by~\cite{HFHPRW2009} containing 117 seconds of EEG eye state corpus with a total of $14,980$ EEG measurements from $14$ different sensors taken with the Emotiv EEG neuroheadset to predict eye states~\cite{SWB2019}. Their dataset is now frequently used as a benchmark and is available on Machine Learning Repository, University of California, Irvine (UCI)~\cite{AN2007}. The reader is referred to~\cite{WGMT2014} for additional information regarding the experiment. 

Following the method in ~\cite{SWB2019}, the data is separated in a training set composed of $80\%$ of the data and a testing set with the remainder. The eye state is encoded as follows: $1$ indicates the eye-closed and $0$ the eye-open state. 

Some statistical analysis was performed on training dataset to check if the data have missing values or any outlier to be fixed. In this respect, were found values corresponding to inaccurate or corrupt records in all-time provided from sensors. The detected inaccurate values are replaced with the mean value of the remaining measurements for each variable. Also, each input sequence is transformed using scale and centering transformations. The Logistic NARX based on FROLS was not able to achieve satisfactory performance when trained with the original dataset. The authors explained the lousy performance as a consequence of the high variability and dependency between the variables measured. Table \ref{tab:metac-egg1} reports that the Meta-MSS$_c$, on the other hand, was capable of building a model with $10$ terms and accuracy of $65.04\%$.  

\begin{table*}[!htb]
	\centering
	\caption{Identified NARX model using Meta-MSS. This model was built using the original EEG measurements. No comparison was made because the FROLS based technique was not capable to generate a model which performed well enough}
	\label{tab:metac-egg1}
		\begin{tabular}{c|ccccccccccc}
			\multirow{2}{*}{Meta-MSS$_c$} & Model term & constant & $x_{\indices{_1}k-1}$ & $x_{\indices{_4}k-30}$ & $x_{\indices{_4}k-36}$ & $x_{\indices{_4}k-38}$ & $x_{\indices{_4}k-41}$ & $x_{\indices{_6}k-2}$ & $x_{\indices{_7}k-5}$ & $x_{\indices{_{12}}k-1}$ & $x_{\indices{_{13}}k-1}$ \\ \cline{2-12} 
			& Parameter  & $0.2055$ & $-0.1077$             & $0.1689$               & $0.1061$               & $0.0751$               & $0.1393$               & $0.3573$              & $-0.7471$             & $-0.4736$              & $0.3875$              
		\end{tabular}%
\end{table*}

This result may appear to be a poor performance. However, the Logistic NARX achieved $0.7199$, and the higher score achieved by the popular techniques was $0.6473$, considering the case where a principal component analysis (PCA) drastically reduced the data dimensionality. Thus, this result shows a powerful performance of the Meta-MSS$_c$ algorithm. For comparison purpose, a PCA is performed, and the first five principal components were selected as a representation of the original data. Table~\ref{tab-metac-comp3} illustrates that Meta-MSSc has built the model with the best accuracy together with the Logistic NARX approach. The models built without autoregressive inputs have the worst classification accuracy, although this is improved with the addition of autoregressive terms. However, even with autoregressive information, the popular techniques do not achieve a classification accuracy to take up the ones obtained by the Meta-MSS$_c$ and Logistic NARX methods.

\begin{table*}[!htb]
	\centering
	\caption{Accuracy performance between different methods for Electroencephalography Eye State Identification}
	\label{tab-metac-comp3}
	\begin{tabular}{cc}
		\hline
		Method                                                 & Classification accuracy \\ \hline
		Meta-MSSc                                              & $0.7480$                \\
		Logistic NARX                                          & $0.7199$                \\
		Regression NARX                                        & $0.6643$                \\
		Random Forest (without autoregressive inputs)          & $0.5475$                \\
		Support Vector Machine (without autoregressive inputs) & $0.6029$                \\
		K-Nearest Neighbors (without autoregressive inputs)    & $0.5041$                \\
		Random Forest (with autoregressive inputs)             & $0.6365$                \\
		Support Vector Machine (with autoregressive inputs)    & $0.6473$                 \\
		K-Nearest Neighbors (with autoregressive inputs)       & $0.5662$                \\ \hline
	\end{tabular}%
\end{table*}

\section{Conclusion}

This study presents the structure selection of polynomial NARX models using a hybrid and binary Particle Swarm Optimization and Gravitational Search Algorithm. The selection procedure considers the individual importance of each regressor along with the free-run-simulation performance to apply a penalty function in candidates solutions. The technique, called Meta-MSS algorithm in its standard form, is extended and analyzed into two main categories: (i) regression approach and (ii), the identification of systems with binary responses using a logistic approach. 

The technique, called Meta-MSS algorithm, outperformed or at least was compatible with classical approaches like FROLS, and modern techniques such as RaMSS, C-RaMSS, RJMCMC, and a meta-heuristic based algorithm. This statement considers the results obtained in the model selection of $6$ simulated models taken from literature, and the performance on the F-16 Ground Vibration benchmark.

The latter category proves the robust performance of the technique using an adapted algorithm, called Meta-MSS$_c$, to build models to predict binary outcomes in classification problems. Again, the proposed algorithm outperformed or at least was compatible with popular techniques such as K-Nearest Neighbors, Random Forests and Support Vector Machine, and recent approaches based on FROLS algorithm using NARX models. Besides the simulated example, the electroencephalography eye state identification proved that the Meta-MSS$_c$ algorithm could handle the problem better than all of the compared techniques. In this case study, the new algorithm returned a model with satisfactory performance even when the data dimensionality was not transformed using data reduction techniques, which was not possible with the algorithms used for comparisons purposes. 

Furthermore, although the stochastic nature of the Meta-MSS algorithm, the individual evaluation of the regressors and the penalty function results in fast convergence. In this respect, the computational efficiency is better or at least consistent with other stochastic procedures, such as RaMSS, C-RaMSS, RJMCMC. The computational effort relies on the number of search agents, the maximum number of iterations, and the search space dimensionality. Therefore, in some cases, the elapsed time of the Meta-MSS is compatible with the FROLS.

The development of a meta-heuristic based algorithm for model selection such as the Meta-MSS permits a broad exploration in the field of system identification. Although some analysis are out of scope and, therefore, are not addressed in this paper, future work are open for research regarding the inclusion of noise process terms in model structure selection, which is an important problem concerning the identification of polynomial autoregressive models. In this respect, an exciting continuation of this work would be to implement an extended version of Meta-MSS to return NARMAX models.


%
%

\bibliographystyle{spphys}       
\bibliography{spnc}   

\end{document}